\documentclass[10pt,twocolumn,letterpaper]{article}

\usepackage{cvpr}              
\usepackage[accsupp]{axessibility} 
\usepackage{multirow}
\definecolor{cvprblue}{rgb}{0.21,0.49,0.74}
\usepackage[pagebackref,breaklinks,colorlinks,allcolors=cvprblue]{hyperref}

\def\paperID{4} 
\def\confName{CVPR}
\def\confYear{2025}

\title{D-Feat Occlusions:  Diffusion Features for Robustness to Partial Visual Occlusions in Object Recognition}

\author{
\begin{tabular}{c@{\hspace{1.5cm}}c}
Rupayan Mallick & Sibo Dong \\
Department of Computer Science & Department of Computer Science \\
Georgetown University, USA & Georgetown University, USA \\
{\tt\small rupayan.mallick@georgetown.edu} & {\tt\small sd1242@georgetown.edu} \\
\\
Nataniel Ruiz\thanks{This work was done when the author was at Boston University.} & Sarah Adel Bargal \\
Google, USA & Department of Computer Science \\
{\tt\small natanielruiz@google.com} & Georgetown University, USA \\
& {\tt\small sarah.bargal@georgetown.edu} \\
\end{tabular}
\vspace{-1em}
}

\begin{document}
\maketitle

\begin{abstract}
Applications of diffusion models for visual tasks have been quite noteworthy. This paper targets making classification models more robust to occlusions for the task of object recognition by proposing a pipeline that utilizes a frozen diffusion model. Diffusion features have demonstrated success in image generation and image completion while understanding image context. Occlusion can be posed as an image completion problem by deeming the pixels of the occluder to be `missing.' We hypothesize that such features can help hallucinate object visual features behind occluding objects, and hence we propose using them to enable models to become more occlusion robust. 
We design experiments to include input-based augmentations as well as feature-based augmentations. Input-based augmentations involve finetuning on images where the occluder pixels are inpainted, and feature-based augmentations involve augmenting classification features with intermediate diffusion features. We demonstrate that our proposed use of diffusion-based features results in models that are more robust to partial object occlusions for both Transformers and ConvNets on ImageNet with simulated occlusions. We also propose a dataset that encompasses real-world occlusions and demonstrate that our method is more robust to partial object occlusions.
\end{abstract}

\vspace{-2.5em}
\section{Introduction}
Occlusions consistently pose challenges for various computer vision systems including medical imaging applications, autonomous vehicles, tracking, object recognition, \textit{etc.} In fact, it has been shown that a prediction for an image can be disrupted by adding human-imperceptible perturbation to the input image \cite{goodfellow2014explaining}, which is equivalent to occluding a small number of image pixels. Deep learning systems are especially affected by occlusions, as they typically focus on learning most discriminative cues; \eg the ear of a dog, in order to classify dog breed. This has been demonstrated in the literature using saliency maps of visual explainability techniques (\eg \cite{zhang2018top}). If such evidence is then occluded, the robustness of the system is greatly affected.

\begin{figure*}[t]
    \centering
    \includegraphics[width=1.0\linewidth]{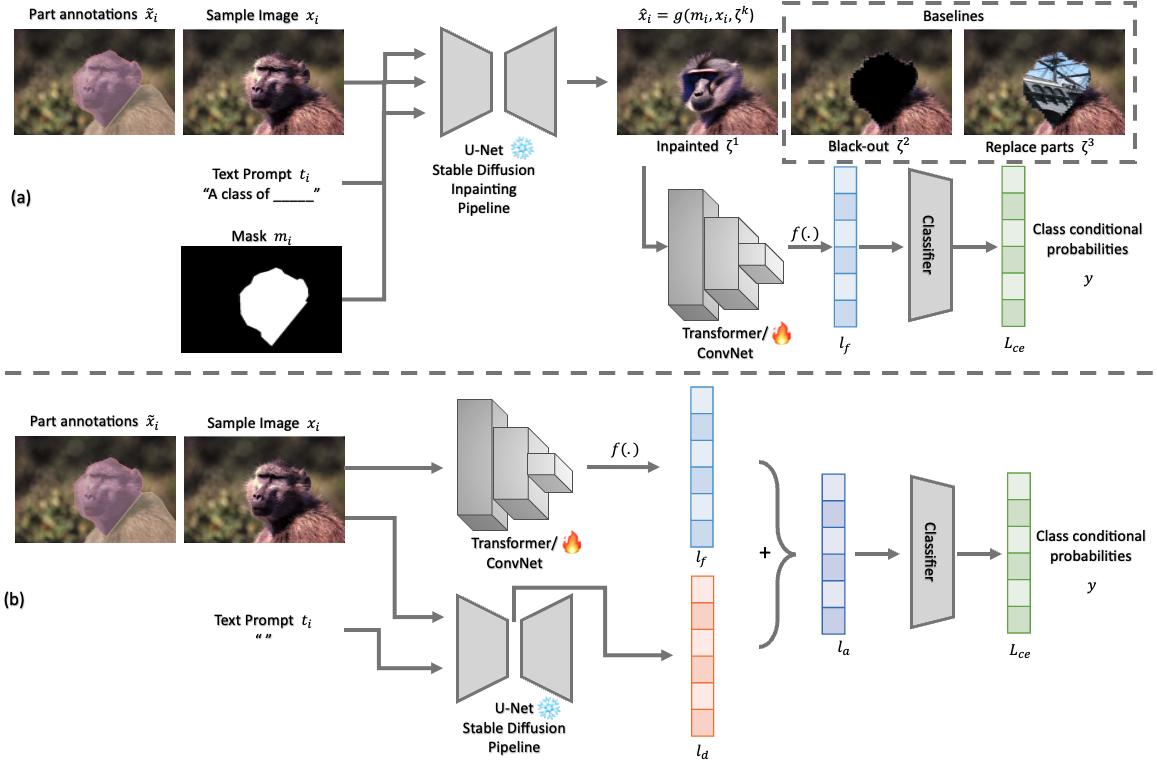}
    \caption{This figure presents the pipeline of our training process for diffusion (a) inpainting augmentation, and (b) feature augmentation. In (a), an input image $x_i$, together with its part annotations $\tilde{x}_i$, simple class text prompt $t_i$, and part segmentation mask $m_i$ are fed into a SD inpainting module that generates an inpainted image $\hat{x}_i$. $\hat{x}_i$ (or baseline image) is fed into the Transformer/ConvNet model to produce model features $l_f$ that are then used for object recognition. In (b), an input image $x_i$, together with its part annotations $\tilde{x}_i$, and null prompt $t_i$ are fed into a frozen SD model where U-Net intermediate features $l_d$ are extracted. $l_f$ and $l_d$ are fused and used as an augmented feature $l_a$ that is then used for object recognition.}
    \label{fig:pipeline}
\end{figure*}

While reasoning about objects and their spatial relations has been addressed for simple 3D geometrical shapes \cite{liu2021learning}, this is significantly more constrained and prohibitive in real-world scenes. As a result, the effect of occlusions on the robustness of models like those of object recognition is significantly underexplored. One main reason for this is the scarcity of real-world datasets in the literature that have real occlusions and not out-of-context objects pasted randomly as distractors. Recently, the large-scale PartImageNet \cite{he2021partimagenet} dataset has been proposed. This dataset provides pixel part-segmentations that are mutually exclusive for single-object images as demonstrated in Figure~\ref{fig:pipeline}. We find that this dataset is instrumental in exploring and providing explanations for the long-standing challenge of occlusion. We use this to develop diffusion-based augmentation techniques, in both the input and feature spaces, that enable the training of models that are more robust to partial visual occlusions. 
Specifically, input space augmentation involves inpainting the candidate image using a diffusion model to hallucinate the missing pixels.
For feature space augmentation, we leverage the diffusion model’s compact yet informative intermediate U-Net features \cite{Rombach_2022_CVPR} with a null prompt.
In doing so, we answer the question: 
\textit{Would augmenting with synthetic image completion or introducing generative features strengthen robustness to occlusions?}

\begin{figure*}[!t]
    \centering
    {\rotatebox[origin=c]{90}{\parbox{1cm}{\centering\textit{Original} }\hspace*{-6em}}} \hspace{0em}
    \includegraphics[width=0.145\linewidth,height=0.145\linewidth,trim={0cm 0cm 0cm 0cm},clip]{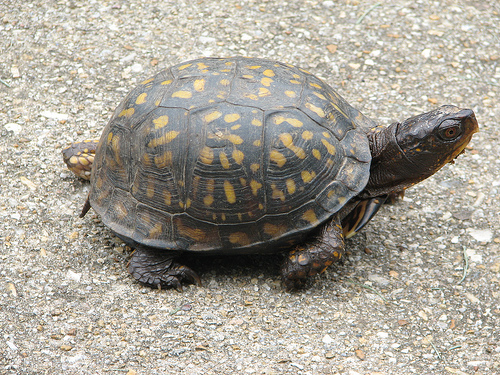} 
    \hspace*{0.1em}
    \includegraphics[width=0.145\linewidth,height=0.145\linewidth,trim={0cm 0cm 0cm 0cm},clip]{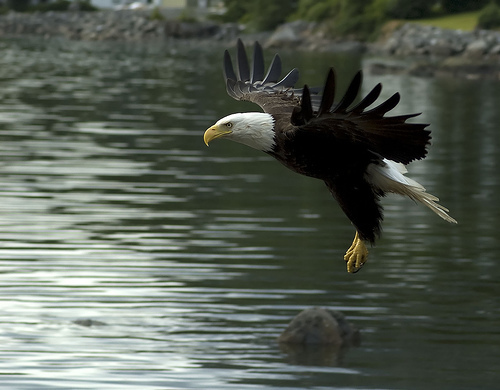} 
    \hspace*{0.1em}
    \includegraphics[width=0.145\linewidth,height=0.145\linewidth,trim={0cm 0cm 0cm 0cm},clip]{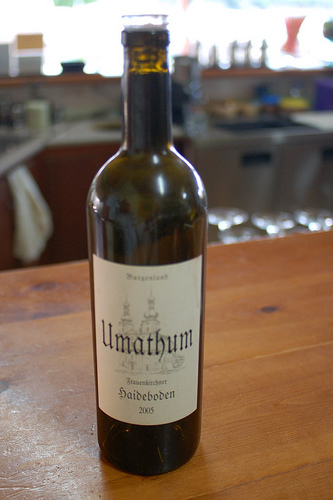}
    \hspace*{0.1em}
    \includegraphics[width=0.145\linewidth,height=0.145\linewidth,trim={0cm 0cm 0cm 0cm},clip]{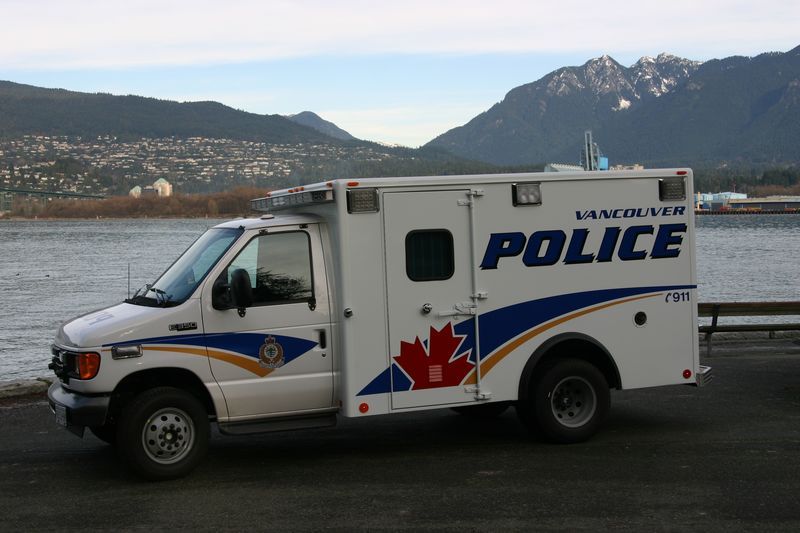}
    \hspace*{0.1em}
    \includegraphics[width=0.145\linewidth,height=0.145\linewidth,trim={0cm 0cm 0cm 0cm},clip]{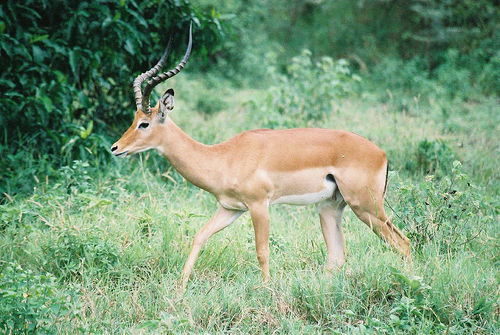}
    \hspace*{0.1em}
    \includegraphics[width=0.145\linewidth,height=0.145\linewidth,trim={0cm 0cm 0cm 0cm},clip]{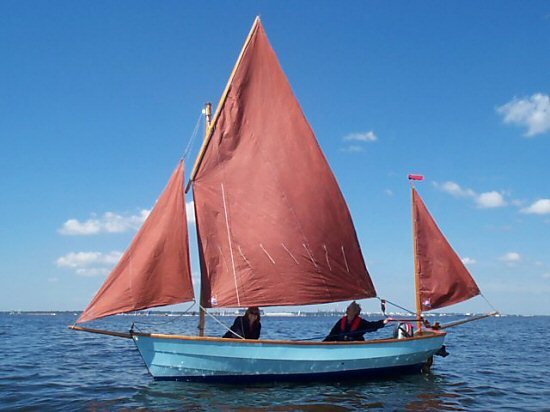}\\[0.2em]
    
    {\rotatebox[origin=c]{90}{\parbox{1cm}{\centering\textit{Occluded} }\hspace*{-6em}}} \hspace{0.2em}
    \includegraphics[width=0.145\linewidth,height=0.145\linewidth,trim={0cm 0cm 0cm 0cm},clip]{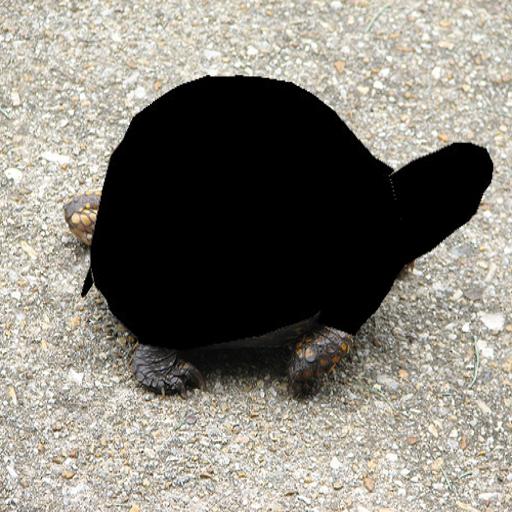} 
    \hspace*{0.1em}
    \includegraphics[width=0.145\linewidth,height=0.145\linewidth,trim={0cm 0cm 0cm 0cm},clip]{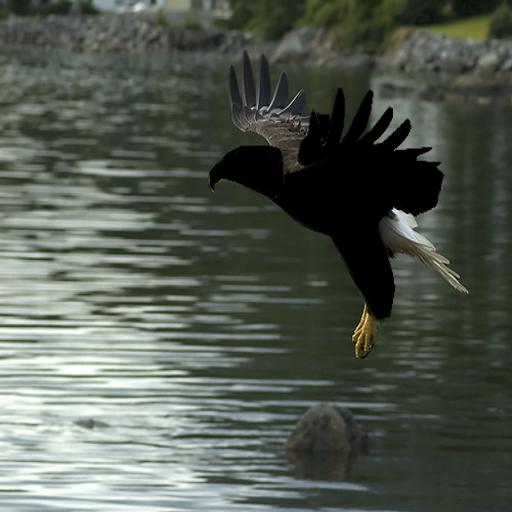} 
    \hspace*{0.1em}
    \includegraphics[width=0.145\linewidth,height=0.145\linewidth,trim={0cm 0cm 0cm 0cm},clip]{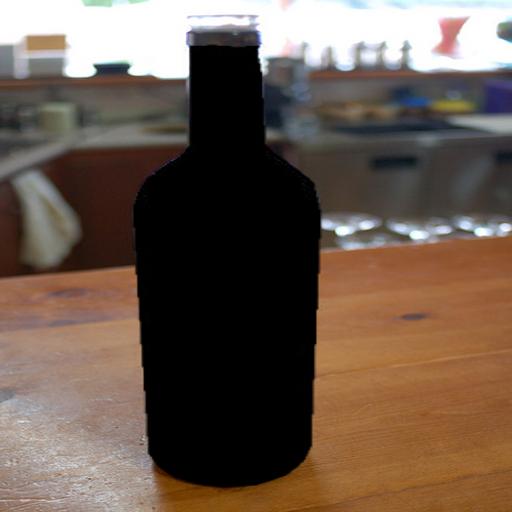}
    \hspace*{0.1em}
    \includegraphics[width=0.145\linewidth,height=0.145\linewidth,trim={0cm 0cm 0cm 0cm},clip]{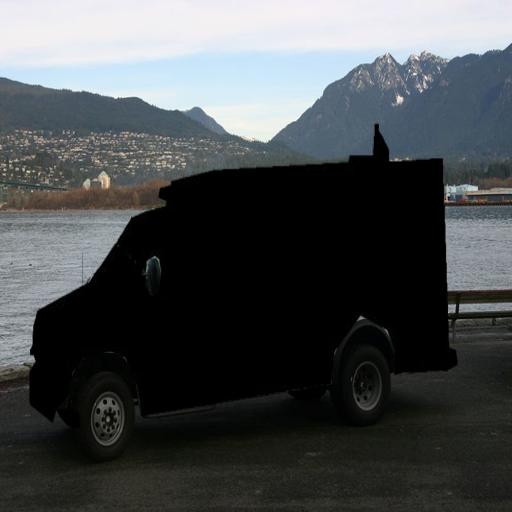}
    \hspace*{0.1em}
    \includegraphics[width=0.145\linewidth,height=0.145\linewidth,trim={0cm 0cm 0cm 0cm},clip]{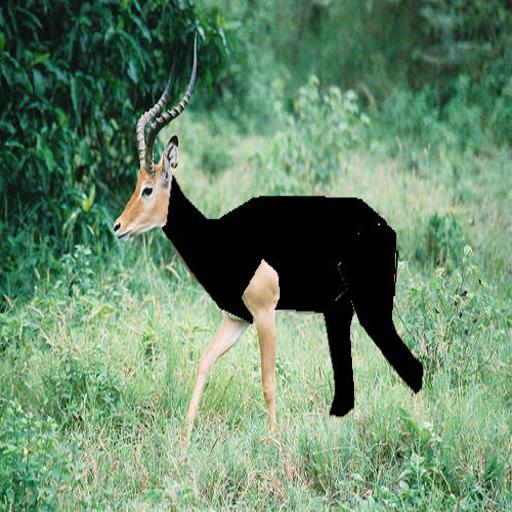}
    \hspace*{0.1em}
    \includegraphics[width=0.145\linewidth,height=0.145\linewidth,trim={0cm 0cm 0cm 0cm},clip]{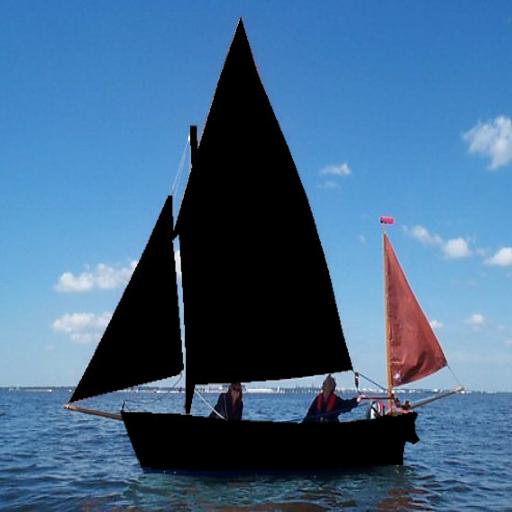}\\[0.2em]
    
    {\rotatebox[origin=c]{90}{\parbox{1cm}{\centering\textit{Inpainted}}\hspace*{-6em}}} \hspace{0em}
    \includegraphics[width=0.145\linewidth,height=0.145\linewidth,trim={0cm 0cm 0cm 0cm},clip]{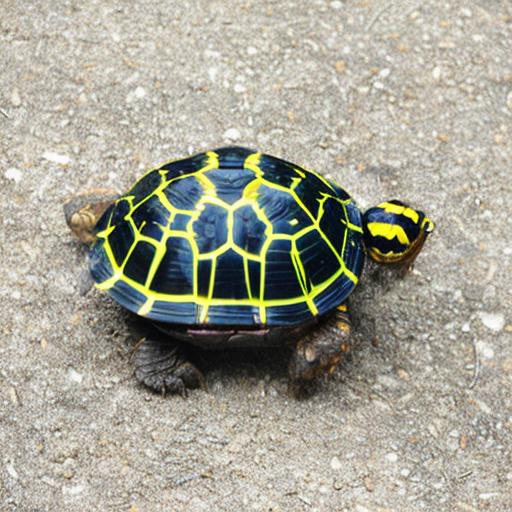} 
    \hspace*{0.1em}
    \includegraphics[width=0.145\linewidth,height=0.145\linewidth,trim={0cm 0cm 0cm 0cm},clip]{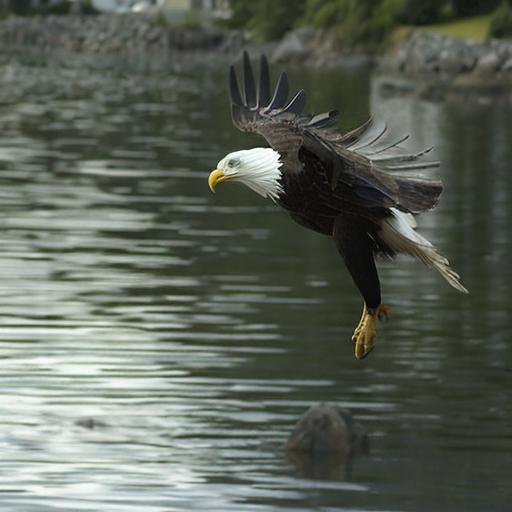} 
    \hspace*{0.1em}
    \includegraphics[width=0.145\linewidth,height=0.145\linewidth,trim={0cm 0cm 0cm 0cm},clip]{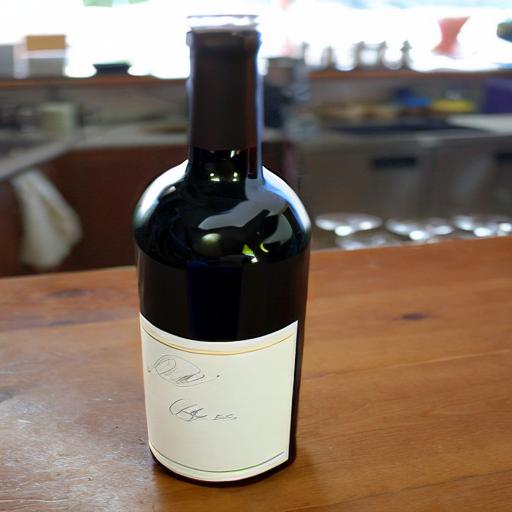}
    \hspace*{0.1em}
    \includegraphics[width=0.145\linewidth,height=0.145\linewidth,trim={0cm 0cm 0cm 0cm},clip]{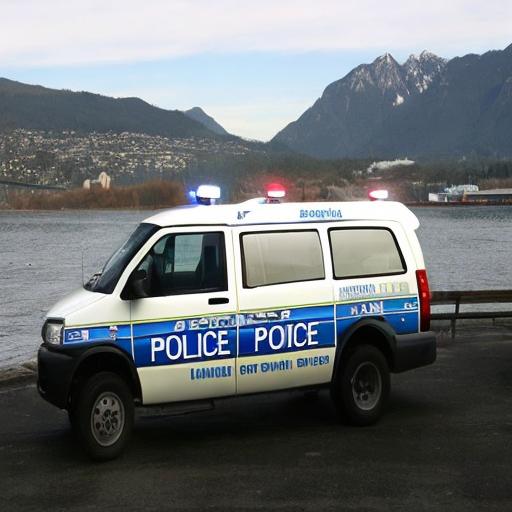}
    \hspace*{0.1em}
    \includegraphics[width=0.145\linewidth,height=0.145\linewidth,trim={0cm 0cm 0cm 0cm},clip]{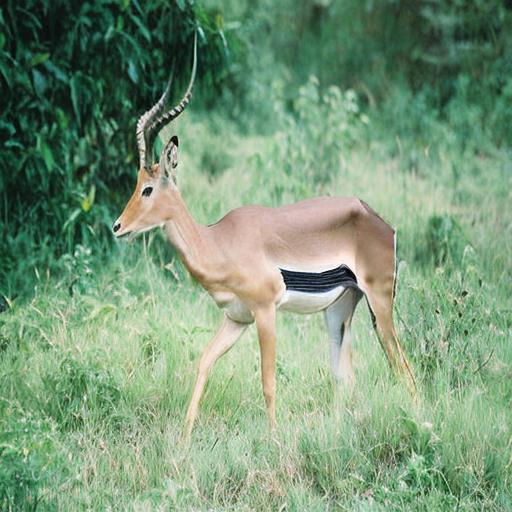}
    \hspace*{0.1em}
    \includegraphics[width=0.145\linewidth,height=0.145\linewidth,trim={0cm 0cm 0cm 0cm},clip]{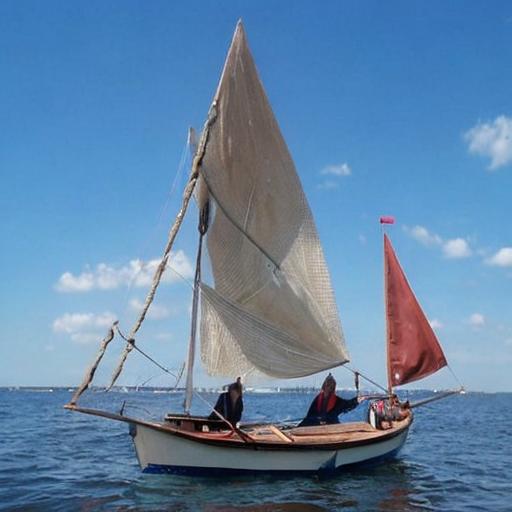}\\[0.2em]

\caption{\textit{Original} (row 1) shows images from the ImageNet training set. \textit{Occluded} (row 2) presents the same images with blacked out image parts mimicking partial object occlusions. \textit{Inpainted} (row 3) presents the generated parts using a Stable Diffusion pipeline \cite{Rombach_2022_CVPR}. The generated inpainted images are then used for input augmentation.}
\label{fig:DA_4}
\end{figure*}

Diffusion models have been recently shown to achieve state-of-the-art results on vision tasks such as image in-painting \cite{Lugmayr_2022_CVPR,Rombach_2022_CVPR}, de-noising \cite{NEURIPS2020_ddpm}, image/video-generation \cite{herzig2019canonical, ho2022imagen}, \etc. Such diffusion models are generative, and therefore are able to generate missing/noisy image pixels. Figure \ref{fig:DA_4} demonstrates sample generation/hallucination of missing pixels resembling occlusion. We hypothesize that features of such models would have the power to encode occluded pixels in the same way missing/noisy pixels are encoded for, and therefore have the capability of hallucinating what is behind occluding objects and help improve robustness of models \wrt occlusions. This is the first work that explores utilizing diffusion features for occlusion robustness. In addition, occlusion is typically evaluated on benchmark datasets of simulated occlusion images. In this work, we propose a real-world occlusion dataset, D-Feat, that has real-world images of occluded objects. Experimental results indicate that our proposed data augmentation methods are more robust to partial object occlusions for both Transformers and ConvNets for simulated and real-world occlusions. We summarize our contributions as follows:
\begin{enumerate}
\item We propose an augmentation technique, in the input space, that leads to models that are more robust to visual occlusion of object recognition systems on both Transformer and ConvNet architectures.
\item We propose an augmentation technique, in the embedding space, that studies the effect of diffusion features on the problem of partial visual occlusion. 
\item We perform a comparative study of how Transformer \vs ConvNet architectures benefit from such augmentations with respect to robustness to occlusions.
\item We curate a real-world occlusion dataset, D-feat, for the evaluation of object classification. We perform evaluations on D-feat, in addition to the standard simulated occlusion on ImageNet.
\end{enumerate}

\section{Related Work}

\textbf{Robustness to Occlusions.} While robustness to occlusions for the task of object detection has been well-studied \cite{10.5555/Faster-RCNN,7780460/yolo}, the final goal is to obtain a polygon that outlines non-occluded parts of objects. Other studies that target locating object boundaries including occluded parts of objects are limited to the work of \cite{wang2020robust} and \cite{Zhu2019RobustnessOO} that use context to help determine occluded object regions. Robustness towards occluded samples is studied in CutMix \cite{yun2019cutmix} where accuracy is reported on image center and image boundary occlusions, regardless to where objects are with respect to the occlusions. \cite{zhong2020random} follow a similar approach to CutOut but use random pixel values on the masked region instead, as opposed to the fixed gray, to simulate the occlusion effect. In contrast, we are studying partial visual occlusions on object parts. \cite{CompCnn} and \cite{Kortylewski_2020_WACV} propose compositional CNNs to tackle the problem of partial occlusion through occluder kernels that help disentangle foreground and background pixels. These approaches consist of a sequential pipeline and focus on ConvNet architectures. We propose a simpler and more compact pipeline and study both Transformer and ConvNet architectures. In contrast to all previous work, studies that target hallucinating object visual features behind occluding objects are lacking for the task of object recognition. 

\textbf{Classical Image Augmentation.} Augmentation methods such as CutMix \cite{yun2019cutmix}, MixUp \cite{zhang2018mixup}, and CutOut\cite{devries2017cutout} target obtaining better regional localization, regularization, and therefore better accuracy models. MixUp and CutOut are aimed at improving regularization by dropping randomly selected rectangular image patches. \cite{zhong2020random} propose RandomErasing, where they randomly select a rectangular region in the image and change the pixel values within the selected region. 
Superpixel-Mix~\cite{franchi2021robust} proposes a semantic-preserving augmentation method that replaces superpixel regions between image pairs to improve robustness in semantic segmentation models. For object detection, \cite{Wang2019DataAF} propose a data augmentation technique in which a pair of images from the same class are used to generate new training instances by interchanging the instances keeping in mind shape and scale information. 
Inspired by early work, we drop image regions; however, in contrast, these regions are part-segmentations obtained using part-annotations and therefore coincide with the object of interest.  Methods such as DropBlock \cite{NEURIPS2018_dropblock} use regional dropout but in the feature space. Because such dropping occurs multiple times when images re-appear in training batches, this can also be viewed as an augmentation technique. In contrast, we augment our model with diffusion features that we hypothesize encode pixels missing due to occlusion.

\textbf{Diffusion Models and Image Augmentation.} Diffusion models have been recently shown to achieve state-of-the-art results on vision tasks such as image in-painting \cite{Lugmayr_2022_CVPR,Rombach_2022_CVPR}, de-noising \cite{NEURIPS2020_ddpm}, image/video-generation \cite{herzig2019canonical, ho2022imagen}, semantic-segmentation \cite{baranchuk2022labelefficient, dpss}, anomaly detection\cite{bandara_anomaly, Wyatt_2022_CVPR}, and other applications \cite{croitoru2023diffusion}. In this work, we propose using the generative power of such features in order to gain the ability of hallucinating visual features that are occluded, and therefore allowing a model to be more robust to occlusions for the task of object recognition.

Recently, diffusion models have also emerged as a novel approach to data augmentation, aimed at enhancing classification performance. These models utilize text-to-image generative techniques to generate images based on carefully engineered prompts of classes \cite{azizi2023synthetic, bansal2023leaving, he2023is, li2023synthetic, marrie2024good, sariyildiz2023fake, shin2023fill, yuan2022not} or captions \cite{dunlap2023diversify}. Several strategies have been developed to increase the diversity of augmentation. \cite{shipard2023diversity} provides a bag of tricks for enhancing diversity during latent diffusion image generation. \cite{zhang2023expanding} introduces noise into the image latent space and applies a diversity loss during optimization. \cite{zhou2023training} implements textual inversion techniques and perturbs the conditioning embedding to generate variations. \cite{fu2024dreamda} modifies the U-Net bottleneck features of the diffusion models during each reverse diffusion step. \cite{trabucco2024effective} adapts diffusion models to new domains by incorporating new concept tokens into the text encoder. \cite{islam2024diffusemix} uses InstructPix2Pix \cite{Brooks_2023_CVPR} for generating multiple versions of the same image and finally generating an augmented image by combining the generated image with the original image.
\cite{shivashankar2023semantic} employs techniques such as latent space perturbation, new concept insertion, and modifications to foreground and background appearances to create alternative yet semantically consistent images. \cite{buburuzan2025mobi} proposes a multimodal object diffusion-based inpainting method, where the model reconstructs occluded object regions using both image and textual context. 

Unlike approaches that rely on diffusion models for generating entire synthetic images, our method introduces two novel uses of diffusion models. First, we employ diffusion-based inpainting to reconstruct missing/occluded image regions, enhancing robustness to partial occlusions. Second, we extract informative diffusion features that enable the model to infer object structures beyond occlusions.

\section{D-Feat Occlusions: Diffusion Features for Robustness to Partial Visual Occlusions}
\label{sec:methods}

In this work, we propose \textit{D-Feat}: input and feature augmentations that utilize part-annotations, and the generative power of diffusion models, in order to make object recognition models more robust \wrt partial object occlusions. 

\textbf{Input Augmentation.} Since part-annotation datasets have been recently proposed \cite{he2021partimagenet}, we use such part-annotations in this work to occlude parts of an object in an image. This occluded image is now a candidate for augmentation at training time. We then inpaint this candidate image using a diffusion model that hallucinates missing pixels, providing a homogeneous image with contextual objects for augmentation. We do so using Stable Diffusion Inpainting \cite{Rombach_2022_CVPR}. We use the ground-truth class of the training data as the text prompt.  

 An input image $x_i$ where $x_i \in \mathbb{R}^{C \times H \times W}$ over the predicted label $y_i$ using a model $f$ can be written as $y_i = f(x_i)$. The augmented image $\hat{x_i}$ is created using the segmented map $m_i$ and augmentation method $\zeta^k$, giving $\hat{x_i} = g(x_i, m_i, \zeta^k)$ where $m_i \in {\{0, 1\}}^{n \times H \times W}$ \ie $n$ binary masks for $n$ parts. Therefore, the neural network $f$ is trained to generalize to a distribution of real and augmented images $Y = f(X, \hat{X})$ where $X= \{x_i\}$, $\hat{X}= \{\hat{x}_i\}$ and $Y= \{y_i\}$ , $i \in \{1, ..., s\}$ where $s$ is the number of instances in the dataset. 

A combination of part-segmentations are used for obtaining the partially occluded images $\hat{X}$. We use half the total number of part-segmentations available for an image, rounded up, so for an object with $n$ parts we use $n/2$ or $(n+1)/2$ parts depending on whether $n \mid 2$ or $n \nmid 2$, respectively. This gives a total number of $n/2$ or $(n+1)/2$ occluded parts for each image. Out of the possible part selections, we choose the part combinations that result in the maximum area of occlusion for the particular image.  The selected image regions are then inpainted $\zeta^1$ (or processed in some other way $\zeta^k$, where $k \in \{2,3,\dots, b\}$ is a specific method from a pool of $b$ baseline methods).

\textbf{Feature Augmentation.} We propose the use of intermediate diffusion features to augment the primary classification features. Diffusion features typically encode information that helps in re-generating an image, sometimes filling in missing/noisy pixels, and we therefore hypothesize that such features would result in added robustness to occlusions when used for training. The intuition behind using the diffusion features is due to its generative ability that can hallucinate the occluded regions. This generative ability is well-studied in the application of inpainting. 
To obtain diffusion features we use Stable Diffusion's compact yet informative intermediate U-Net features \cite{Rombach_2022_CVPR} with a null prompt. 

\begin{figure*}[!t]
    \centering 
    {\rotatebox[origin=c]{90}{\parbox{1cm}{\centering\textit{{ImageNet}} \\ \textit{Validation}}\hspace*{-8em}}} \hspace{0.2em}
    \includegraphics[width=0.30\linewidth,height=0.23\linewidth,trim={0cm 0cm 0cm 3cm},clip]{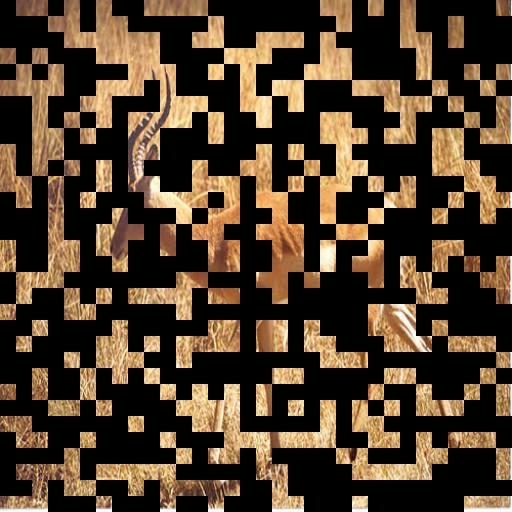} 
    \hspace*{0.1em}
    \includegraphics[width=0.30\linewidth,height=0.23\linewidth,trim={0cm 0cm 0cm 3cm},clip]{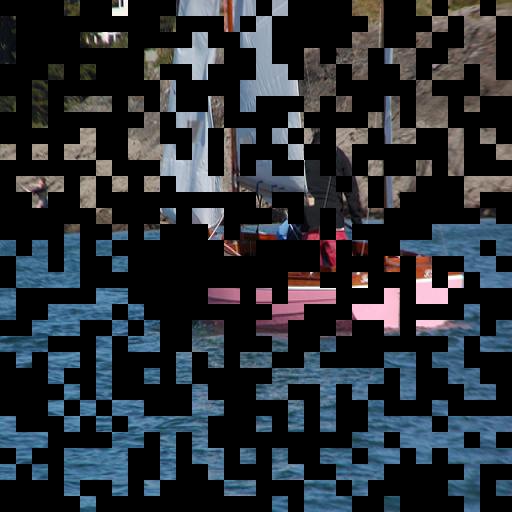} 
    \hspace*{0.1em}
    \includegraphics[width=0.30\linewidth,height=0.23\linewidth,trim={0cm 0cm 0cm 3cm},clip]{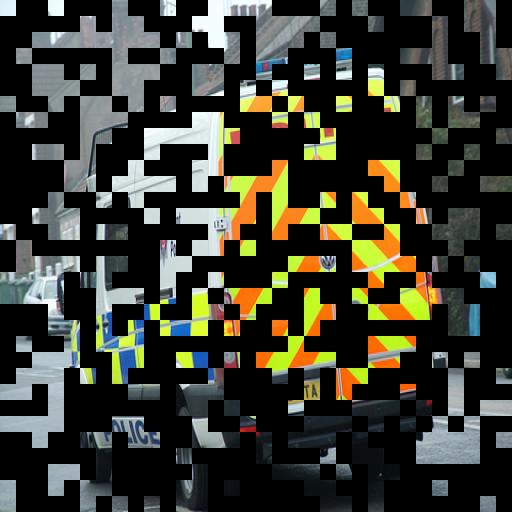}\\[0.2em]
    {\rotatebox[origin=c]{90}{\parbox{1cm}{\centering\textit{Real} \\ \textit{Images}}\hspace*{-8em}}} \hspace{0.2em}
    \includegraphics[width=0.30\linewidth,height=0.23\linewidth,trim={0cm 0cm 0cm 3cm},clip]{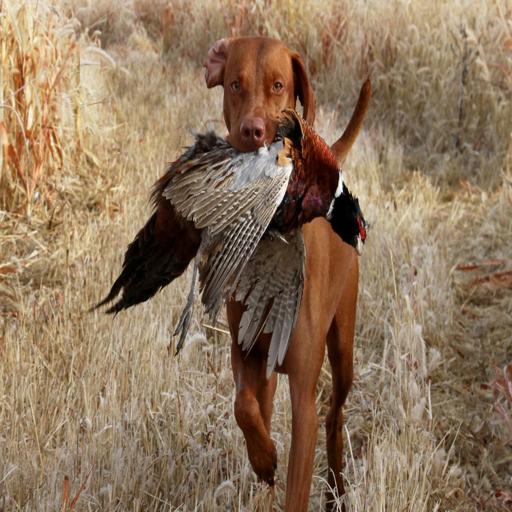} 
    \hspace*{0.1em}
    \includegraphics[width=0.30\linewidth,height=0.23\linewidth,trim={0cm 0cm 0cm 3cm},clip]{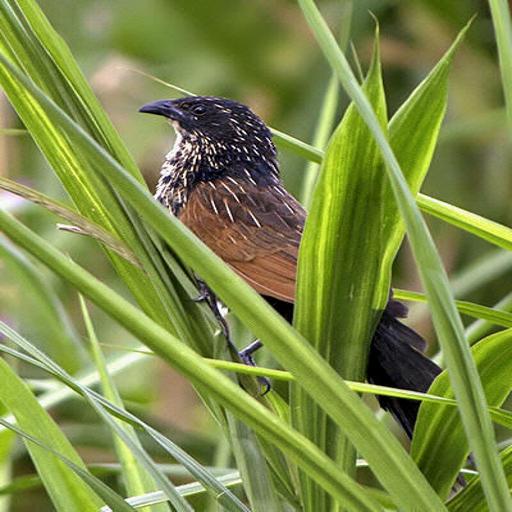} 
    \hspace*{0.1em}
    \includegraphics[width=0.30\linewidth,height=0.23\linewidth,trim={0cm 0cm 0cm 3cm},clip]{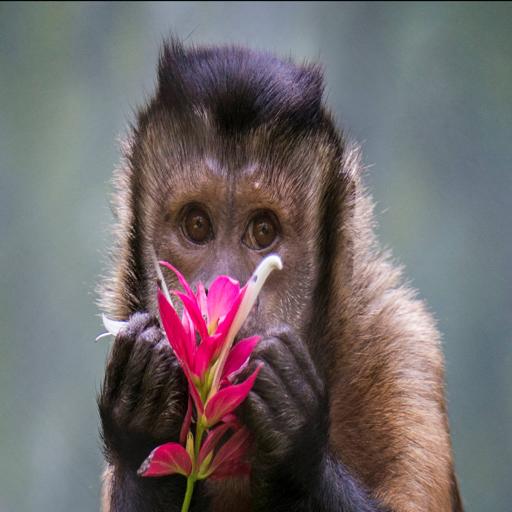}\\[0.2em]
\caption{This figure contrasts the two test setups we use. 
Row 1: Occlusions in the ImageNet Validation \cite{imagenet_cvpr09} set for a sample $60\%$ occlusion. Row 2: Real-world images from our D-feat dataset crawled to demonstrate occlusion of objects from particular classes of interest that overlap with the PartImageNet dataset.}
\label{fig:DA_3}
\end{figure*}






\textbf{Baselines.} Because of the scarcity of part-labeled data, previous baselines mostly relied on rectangular or a standardized-shape occlusion. Therefore, to study the effect of our proposed diffusion based augmentations, we need to propose more reasonable baselines for comparison that also train on part-labeled data. We design three such basic augmentation baselines: $\zeta^1$, $\zeta^2$, and $\zeta^3$. 

$\zeta^1$\textit{: blacking out the parts.} This studies robustness of the model when texture and color are eliminated from regions defined by $m_i$, and only shape information remains as presented in Figure~\ref{fig:DA_4}. $\zeta^2$\textit{: using CutMix \cite{yun2019cutmix}}. This uses the CutMix algorithm to compute the robustness of the method. This method is not governed by the $m_{i}$ but rather a dropout probability. $\zeta^3$\textit{: replacing parts using another image.}
Regions defined by $m_i$ are replaced with pixels from another image. Together with segmentation maps, these enhance the training images. The intuition is to ensure the network focuses on the shape of the object as the texture has been replaced. We propose a loss that classifies both the augmented set of images and the mask. This is presented in Equation~\ref{eq:comb_loss}, where $\alpha$ is the weighting parameter for classifying the augmented set of images and $\beta$ is the weighting parameter for classifying the masks.

\begin{equation}\label{eq:comb_loss}
\begin{aligned}
    \underset{\theta}{\mathrm{arg\,min}}  \Big[ &\alpha \mathbb{E}_{(\{x_{i}, \hat{x}_{i} \}, y_{i})} \left[ {L}(\theta, \{x_{i}, \hat{x}_{i}\}, y_{i}) \right] \\
    + &\beta \mathbb{E}_{(m_{i}, y_{i})} \left[ {L}(\theta, m_{i}, y_{i}) \right] \Big]
\end{aligned}
\end{equation}

\section{Experiments}

In this section, we first discuss dataset choices for training and evaluation. Next, we present our experimental setup summarizing architectures and hyper-parameters used for reproducibility purposes. We then present our results, discuss how the proposed augmentation techniques improve model robustness to occlusions.

\subsection{Datasets} 

In this work, we use the PartImageNet \cite{he2021partimagenet} dataset 
and the ImageNet1K Validation (Val) dataset. In addition, we curate a real-world occlusion dataset, D-feat. D-feat was crawled to demonstrate real occlusion of objects. We crawl three real images that demonstrate real occlusion for each of 100 classes of the PartImageNet \cite{he2021partimagenet} dataset. Curating images for the remaining 58 classes of PartImageNet was found to be challenging due to their correspondence with objects that are not commonly observed to be occluded in the real world. For example, it is difficult to find real images of occluded airliners.

\textbf{Training.} PartImageNet is used for the training of our models. It consists of 24,095 images of ImageNet \cite{imagenet_cvpr09} (158 classes) and their corresponding part annotations (COCO-style). This dataset has been primarily used for object detection and has disjoint classes in the training, validation, and test sets. For our task we use it for training on the classification task of object recognition, and therefore validation requires all the classes to be present. While other datasets with part annotations exist, such as PACO \cite{ramanathan2023paco} which contains images from the COCO 2017 dataset \cite{Lin2014MicrosoftCC}, a single image can have multiple object instances. Therefore, directly using it for image classification and interpreting robustness to occlusions can be challenging. Other datasets that have part annotations include PASCAL-Part \cite{Chen_2014_CVPR} and Cityscapes \cite{Cordts2016Cityscapes}, these again have multiple objects per image and/or have a small number of classes compared to PartImageNet.

\begin{table*}[t]
  \caption{Occlusion study on the $158$ intersection classes of the ImageNet1K Val dataset. The highest top-1/top-5 accuracy for each information loss column (marked in bold) lies in one of the two proposed input/embedding space augmentations with larger performance gains associated with higher information loss. While \textit{SD Features} uses a pre-trained SD model to extract features for each of our training/testing images, \textit{SD Inpaint} inpaints occluded training images and uses the inpainted images to augment the training set. CutMix can be seen performing similar to our proposed methods using the features from stable diffusion on this simulated dataset, but this might be due to how the training is performed where the ground truth labels are presented propotionally to that of area of the patches.
    } \vspace*{1em}
    \label{tab:part-imagenet}
  \centering
  \begin{tabular}{r|c|cccccc}
    \toprule
     &  & \multicolumn{5}{c}{\textbf{Accuracy (top-1 / top-5) for Information Loss:}}                   \\
    \textbf{Augmentation Method} & \textbf{Architecture}    & \textbf{0\%}  & \textbf{20\%}  & \textbf{40\%} & \textbf{60\%}  & \textbf{80\%}   \\
    \midrule
    \midrule
    \multirow{2}{*}{\textbf{\textit{None}}} & ConvNeXt & 87.68 / 97.78 & 86.48 / 97.32 & 83.98 / 96.25 & 78.94 / 93.84 & 57.27 / 78.36  \\
    & Swin   & 88.65 / 98.64 & 88.03 / 98.49 & 86.79 / 98.17 & 83.88 / 97.27 & 71.87 / 92.36   \\
    \midrule
    \multirow{2}{*}{\textbf{\textit{Black-Out Parts}}} & ConvNeXt  & 88.01 / 98.16 & 86.69 / 97.82 & 84.70 / 97.05 & 79.51 / 94.93 & 59.32 / 82.26    \\
    & Swin  & 88.78 / 98.65 & 88.06 / 98.53 & 86.77 / 98.26 & 83.78 / 97.37 & 72.77 / 92.83    \\
    \midrule
    \multirow{2}{*}{\textbf{\textit{Replace Parts}}} & ConvNeXt  &  88.15 / 98.45 & 86.21 / 97.82 & 83.48 / 97.07 & 77.49 / 94.36 & 62.87 / 84.62 \\
    & Swin  &  89.01 / 98.62 & 88.03 / 98.38 & 86.34 / 98.02 & 82.90 / 97.07 & 72.41 / 92.68
    \\
    \midrule
    \multirow{2}{*}{\textbf{\textit{CutMix \cite{yun2019cutmix}}}} & ConvNeXt  & 88.12 / 97.68 & 86.96 / 97.38 & 84.90 / 96.73 & 80.91 / 94.95 & 67.44 / 86.90    \\
    & Swin  & 88.75 / 98.63 & 88.24 / 98.50 & 87.17 / 98.27 & \textbf{85.11} / 97.75 & 77.22 / 95.18    \\
    \midrule
    \multirow{2}{*}{\textbf{\textit{Ours: SD Inpaint}}} & ConvNeXt & 88.04 / 98.32 & 87.19 / 97.82 & 85.91 / 97.20 & 81.04 / 95.17  & 69.84 / 89.35  \\
    & Swin   & 89.01 / 98.76 & \textbf{88.68} / \textbf{98.91} & \textbf{87.83} / 98.79 & 84.85 / 97.79 & 77.96 / \textbf{95.92} \\
    \midrule
    \multirow{2}{*}{\textbf{\textit{Ours: SD Features}}} & ConvNeXt & 88.95 / 98.13 & 88.10 / 97.96 & 85.96 / 96.94 & 82.80 / 95.07  & 70.19 / 87.63  \\
    & Swin   &  \textbf{89.07} / \textbf{98.81} & 88.39 / 98.63 & 86.94 / \textbf{98.81} & {85.07} / \textbf{97.82}  & \textbf{78.14} / 95.53 \\
    \midrule
    \bottomrule 
  \end{tabular}
\end{table*}

\textbf{Evaluation: Simulated Occlusion.} 
The ImageNet1K Val dataset is used for the evaluation of our augmentation techniques. Following the partial occlusion work of \cite{CompCnn}, we only consider the intersection of classes of each of these datasets with PartImageNet. To simulate occlusion on test sets, we randomly black-out patches of size $16 \times 16$ from the images. We note that some patches that are close to the image borders may end up having smaller occlusions. 


\textbf{Evaluation: Real Occlusion.} Additionally, we use the D-Feat dataset to test real-world occlusions as opposed to the typically used simulated occlusions. D-feat is used to evaluate and compare augmentation techniques. Figure ~\ref{fig:DA_3}, presents samples of simulated and real occlusions.


\subsection{Experimental Setup}

We use standard deep Transformer and ConvNet architectures. For the Transformer model, we use Swin-Base \cite{liu2021Swin}, and for the ConvNet model, we use ConvNeXt-Base \cite{liu2022convnet}. From an experimental point of view, these models have similar number of parameters and FLOPs. The models used for this work are pre-trained on ImageNet1K. We fine-tune these models using the very recently published, largest single-object publicly available part-annotations dataset: PartImageNet \cite{he2021partimagenet}. To evaluate the models, we use the ImageNet Validation dataset, limited to the subset of classes of PartImageNet. PartImageNet is not used for evaluation as it overlaps with the ImageNet training dataset, which has been used to pre-train the models. 

For both Swin and ConvNeXt architectures, we use a cross-entropy classification loss, SGD training, a batch size of 64, a learning rate 10e-3 (selected using the grid search approach), and  trained for 25-30 epochs until convergence for each model. Each experiment is launched using a node of four 16GB Tesla T4 GPUs in parallel and the diffusion features experiments are performed on a single A100 GPU. In total, we use two such nodes to conduct our experiments. For the diffusion model used to perform the inpainting and obtain the features used for augmentation, we use Stable Diffusion (SDv2-1) following the proposed hyperparameters in \cite{Rombach_2022_CVPR}. The text prompt used for inpainting is simplistic ``A class of $<$object\_class$>$'', and the text prompt used for diffusion feature extraction is the Null prompt. A frozen SD model was found to inpaint well enough due to being pre-trained on the large-scale LAION-5B dataset \cite{schuhmann2022laionb}, and therefore fine-tuning was not essential.

\begin{table}[t]
    \caption{Occlusion study on the D-Feat real-world occlusion dataset comprising of 100 classes. Both ConvNeXt and Swin architectures benefit from all the proposed types of augmentations compared to the No Augmentation baseline. Our proposed augmentations give significantly higher classification accuracies compared to other augmentation techniques on real-world occluded data. We note that we did *not* use the D-Feat dataset for training because it does not include part-annotations. 
    } \vspace*{1em}
  \label{tab:crawled_dataset}
  \centering
    \begin{tabular}{l|c|c}
    \toprule
     \textbf{Augmentation Method} &\textbf{ConvNeXt} & \textbf{Swin}\\
     \midrule
     \midrule
    {\textbf{\textit{None}}} & 82.03 & 76.99\\

     \midrule
     {\textbf{\textit{Black-out Parts}}}  & 86.71 & 88.33\\
     \midrule
     {\textbf{\textit{Replace Parts}}}  & 87.89 & 86.33 \\
      \midrule
     {\textbf{\textit{CutMix}}\cite{yun2019cutmix}} & 85.41 &  86.99 \\
     \midrule
    {\textbf{\textit{Ours: SD Inpaint}}} & 87.10 &  \textbf{89.16} \\
     \midrule
     {\textbf{\textit{Ours: SD Features}}} & \textbf{90.06} &  88.08  \\
    \midrule
    \bottomrule
  \end{tabular}
\end{table} 

 \subsection{Results}

We now present our experiments and discuss the obtained results to evaluate the occlusion robustness of object recognition models using our proposed augmentation techniques. Table~\ref{tab:part-imagenet} presents the top-1 and top-5 classification accuracy for the ImageNet1K Val set. We test on various information losses: 20\%, 40\%, 60\%, 80\% for the ImageNet Val dataset. In this table we demonstrate how input space and embedding space augmentations help improve the accuracy of a baseline model with no augmentations for both ConvNeXt and Swin architectures. In Table~\ref{tab:part-imagenet}, \textit{SD Features} use a pre-trained SD model to extract features for each of our training/testing images, and \textit{SD Inpaint} inpaints occluded training images and uses the inpainted images to augment the training set. In Table~\ref{tab:part-imagenet}, the highest top-1/top-5 accuracy for each information loss column lies in one of the two proposed input/embedding space augmentations with larger performance gains associated with higher information loss, \ie more simulated occlusion. Our proposed augmentation techniques demonstrate higher accuracies for most occlusion percentages. We note that the test set here consists of $\sim$8K images, and therefore accuracy increase corresponds to a large number of correct classifications. The results also highlight the robustness of the two tested architectures while increasing the degree of information loss for simulated occlusion.



Table~\ref{tab:crawled_dataset} presents the results for the real-world D-Feat dataset. Our proposed augmentations significantly outperform other augmentation techniques demonstrating a clear benefit of its use in real-world occlusion scenarios.

Previous work \cite{naseer2021intriguing} has presented that transformers are in general robust to the dropping of patches (simulated occlusion) especially when comparing against the traditional ResNet models to DeIT. \cite{ruiz2022finding} studies this on more recent Transformer and ConvNet architectures as the recent ConvNet architectures include design ingredients such as patchifying and layernorm that target bridging the performance gap with Transformer models such as Swin. The main question we address here is: \textit{Do the ingredients of modern Transformer and ConvNet architectures, such as Swin and ConvNeXt, affect robustness to occlusion?} This motivated our selection of ConvNeXt and Swin as base architectures in this robustness to occlusion study. Directly comparing to the traditional ConvNets such as ResNet, VGG, \etc is not ideal as they are extremely non-robust to patch drop and occlusions as stated in \cite{naseer2021intriguing}.

Tables~\ref{tab:part-imagenet} and~\ref{tab:crawled_dataset} show a comprehensive comparison between the ConvNet ConvNeXt and the Transformer Swin with regards to the effectiveness of the augmentation techniques while dealing with occlusions. Table~\ref{tab:part-imagenet} demonstrates apparent higher robustness (higher accuracy) to occlusions for Swin \vs ConvNeXt, however, the baseline accuracy for Swin was initially higher than Swin using a similar accuracy gap. Therefore, both ConvNeXt and Swin architectures seem to have similar robustness to occlusions, and seem to both benefit similarly from all the proposed augmentation techniques. It is evident that diffusion based augmentation methods outperform other augmentation methods for most degrees of occlusion. 
Table~\ref{tab:crawled_dataset} shows similar robustness to occlusions for both ConvNeXt and Swin architectures.

\section{Conclusions}
In this work we propose two strategies for making deep models more robust to occlusions. The first strategy is based on input space augmentations that utilize diffusion-based image inpainting versions of occluded regions of a sample training image. The second strategy is based on diffusion features that are used as embedding space augmentations. Such diffusion features by design integrate information about missing/noisy pixels, and therefore are able to help make models more robust to the occluded pixels. We demonstrate that our proposed augmentation techniques improve model robustness on a standard benchmark dataset that simulates occlusion, with diffusion feature augmentations showing the best performance. We also find that the augmentation of diffusion based features performs similarly to diffusion based inpainting, while being significantly more efficient. Additionally, we curate a real-world occlusion dataset for evaluation on real occlusions. This gives us an opportunity to move beyond the simulated datasets and move towards more naturatulitic occlusions. Diffusion based augmentation methods outperform all prior input augmentation techniques on images possessing real occlusions. We anticipate that using diffusion based features will have a broader impact and play a big role in making models more occlusion robust, enabling better recognition of objects for high stakes applications like autonomous vehicles that need to recognize pedestrians whether or not they may be occluded.


\section{Limitations and Negative Impacts}
\textbf{Limitations. }\label{subsec:limitations}
Simulated occlusions do not resemble real-world occlusions, but do have the benefit of quantification of the occluded regions, \textit{e.g.} determining that an image has 40\% occlusion. Testing on real-world occlusion data has the limitation of non-quantifiable occlusion percentages (although segmentation algorithms with some human input could be employed to determine these). Additionally, there is no control in simulated occlusions on which regions are occluded, while human bias when taking photos might result in less occlusions of highly discriminative regions of an object. Furthermore, our input augmentation method leverages the Stable Diffusion inpainting pipeline, and therefore inherits the limitations and biases present in Stable Diffusion. This may influence the quality of the augmented data.

\textbf{Negative Impacts. }\label{subsec:negative_impacts}
We propose two data augmentation strategies to make deep models more robust to occlusions. However, data augmentation, when applied to biased training data, can inadvertently amplify existing biases resulting in undesired behaviors such as  misclassifying or underperforming on certain minority groups. Moreover, data augmentation increases computational load, as more data needs to be processed during model training. This can lead to a significant increase in energy consumption and carbon emissions. To mitigate this, we utilized frozen weights whenever possible.

{
    \small
    \bibliographystyle{ieeenat_fullname}
    \bibliography{main}
}


\end{document}